\documentclass[letterpaper]{article}
\usepackage{times}
\usepackage{helvet}
\usepackage{courier}

\usepackage{amsmath}
\usepackage{amssymb}
\usepackage{comment}

\newcommand{\shortcite}[1]{\cite{#1}}

\begin{document}
%

\title{On-the-fly Macros}

\author{Hubie Chen\\
Dept. of Information and Communication Technologies\\
Universitat Pompeu Fabra\\
Passeig de Circumval$\cdot$laci\'o, 8\\
08003 Barcelona, Spain\\
hubie.chen@upf.edu
\and
Omer Gim\'enez\\
Dept. of Llenguatges i Sistemes Inform\`atics\\
Universitat Polit\`ecnica de Catalunya\\
Jordi Girona, 1-3\\
08034 Barcelona, Spain\\
omer.gimenez@upc.edu
}

\maketitle
\begin{abstract}
\begin{quote}
We present a domain-independent algorithm that
computes macros in a novel way.
Our algorithm computes macros ``on-the-fly'' for a given set of states
and does not require previously learned or inferred information, nor 
prior domain knowledge.
The algorithm is used to define
new domain-independent tractable classes of classical planning
that are proved  
 to include \emph{Blocksworld-arm} and
\emph{Towers of Hanoi}.
\end{quote}
\end{abstract}


\newtheorem{theorem}	 			{Theorem}
\newtheorem{lemma}		[theorem]	{Lemma}	
\newtheorem{definition}		[theorem]	{Definition} 
\newtheorem{prop}		[theorem]	{Proposition} 

\newtheorem{domaincore}[theorem]{Domain}

\newenvironment{domain}
  {\begin{domaincore}\rm}
  {$\Box$ \end{domaincore}}

\newtheorem{examplecore}[theorem]{Example}

\newenvironment{example}
  {\begin{examplecore}\rm}
  {$\Box$ \end{examplecore}}

\newenvironment{pf}{\noindent\textbf{Proof\/}.}{$\Box$ \vspace{1mm}}
\newenvironment{pf-sketch}{\noindent\textbf{Proof (Sketch)\/}.}{$\Box$ \vspace{1mm}}

\newcommand{\longversion}[1]{}
\newcommand{\proofsversion}[1]{#1}
\newcommand{\shortversion}[1]{#1}
\newcommand{\oldshortversion}[1]{}

\newcommand{\proof}[1]{
{\noindent {\it Proof.} {#1} \rule{2mm}{2mm} \vskip \belowdisplayskip}
}
\newcommand{\n}[1]{
{\| {#1} \|}
}
\newcommand{\subclaim}[1]{
\vskip 0.10in
{\noindent {\bf Claim: } {\em {#1}}}
\vskip 0.10in
}

\newcommand{\prevproof}[3]{
{\noindent {\bf Proof of {#1}~\ref{#2}.} {#3} \rule{2mm}{2mm} \vskip \belowdisplayskip}
}

\def\Z{\mathcal{Z}}
\def\R{\mathcal{R}}
\def\I{\mathcal{I}}
\def\P{{\mathcal{P}}}
\def\eps{\epsilon}
\def\sm{\setminus}
\def\xb{\bar{x}}

\newcommand{\nat}{\mathbb{N}}

\newcommand{\init}{\mathsf{init}}
\newcommand{\goal}{\mathsf{goal}}
\newcommand{\vars}{\mathsf{vars}}
\newcommand{\pre}{\mathsf{pre}}
\newcommand{\post}{\mathsf{post}}
\newcommand{\restrict}{ \upharpoonright }
\newcommand{\wrong}{\mathsf{wrong}}

\newcommand{\bin}{\mathcal{B}}

\newcommand{\ham}{\mathsf{d_h}}

\newcommand{\transitive}{\mathsf{transitive}}
\newcommand{\apply}{\mathsf{apply}}
\newcommand{\better}{\mathsf{better}}
\newcommand{\addlabel}{\mathsf{addlabel}}
\newcommand{\combine}{\mathsf{combine}}

\newcommand{\arm}{\mathsf{arm}}
\newcommand{\btable}{\mathsf{table}}
\renewcommand{\empty}{\mathsf{empty}}
\newcommand{\on}[1]{#1\mathsf{\mbox{-}on}}
\newcommand{\clear}[1]{#1\mathsf{\mbox{-}clear}}
\newcommand{\pickup}{\mathsf{pickup}}
\newcommand{\putdown}{\mathsf{putdown}}
\newcommand{\stack}{\mathsf{stack}}
\newcommand{\unstack}{\mathsf{unstack}}
\newcommand{\subtowerToTable}{\mathsf{subtow\mbox{-}table}}
\newcommand{\subtowerToPosition}{\mathsf{subtow\mbox{-}pos}}
\newcommand{\towerToBlock}{\mathsf{tow\mbox{-}block}}
\newcommand{\subtowerToBlock}{\mathsf{subtow\mbox{-}block}}
\newcommand{\true}{\mathsf{T}}
\newcommand{\false}{\mathsf{F}}

\section{Introduction}

Macros have long been studied in AI planning~\cite{fn71,korf85}.  
Many domain-dependent applications of macros have been exhibited
and studied~\cite{iba89,js01,h01}; also, a number of domain-independent methods
for learning, inferring, filtering, and applying macros have been
the topic of research continuing up to the present~\cite{bems05,cs07,nlfl07}.

In this paper, we present a domain-independent algorithm that
computes macros in a novel way.
Our algorithm computes macros ``on-the-fly'' for a given set of states
and does not require previously learned or inferred information, nor does it
need any prior domain knowledge.
We exhibit the power of our algorithm by using it to define
new domain-independent tractable classes of classical planning that strictly
extend previously defined such classes~\cite{act-local}, 
and can be proved to include \emph{Blocksworld-arm} and
\emph{Towers of Hanoi}.
We believe that this is notable as
theoretically defined, domain-independent
tractable classes have generally struggled to
incorporate construction-type
domains such as these two.
We hence give theoretically grounded evidence of the computational
value of macros in planning.

\paragraph{\bf Our algorithm.}
Consider the following reachability problem:
given an instance of planning and a
set $S$ of states, compute the ordered pairs of states
$(s, t) \in S \times S$ such that the second state $t$ is reachable
from the first state $s$.  (By \emph{reachable}, we mean that
there is a sequence of operators that transforms the first state into
the second.)  This problem is clearly hard in general, as
deciding if one state is reachable from another captures the complexity
of planning itself.

A natural--albeit incomplete--algorithm for solving this reachability problem
is to first compute the pairs $(s, t) \in S \times S$ such that
the state $t$ is reachable from the state $s$ by application of a single
operator, and then to compute the transitive closure of these pairs.
This algorithm is well-known to run in polynomial time 
(in the number of states and the size of the instance)
but will only discover pairs for which the reachability is evidenced
by plans staying within the set of states $S$:
the algorithm is efficient but incomplete.

The algorithm that we introduce is a strict generalization of
this transitive closure algorithm for the described reachability problem.
We now turn to a brief, high-level description of our algorithm.
Our algorithm begins by computing the pairs connected by a single operator,
as in the just-described algorithm, but each pair is labelled with 
its connecting operator.
The algorithm then continually applies two types of transformations to the
current set of pairs until a fixed point is reached.
Throughout the execution of the algorithm, every pair has an associated
label which is either a single operator or a macro derived by
combining existing labels.
The first type of transformation 
(which is similar to the transitive closure)
is to take pairs of states
having the form $(s_1, s_2)$, $(s_2, s_3)$ and to add the
pair $(s_1, s_3)$ whose new label is the macro
obtained by ``concatenating''
the labels of the pairs $(s_1, s_2)$ and $(s_2, s_3)$.
If the pair $(s_1, s_3)$ is already contained in the current set,
the algorithm replaces the label of $(s_1, s_3)$ with the
new label if the new label is ``more general'' than the old one.\footnote{
For the precise definitions of ``concatenation'' and ``more general'',
please refer to the technical sections of the paper.
}
The second type of transformation is to take a state $s \in S$ and
a label of an existing pair, and to see if the label applied to $s$
yields a state $t \in S$; if so, the pair $(s, t)$ is introduced, 
and the same replacement procedure as before is invoked if the pair
$(s, t)$ is already present.

Our algorithm, as with the transitive closure, operates in 
polynomial time
(as proved in the paper) and is incomplete.  
We want to emphasize that it can, in general, identify pairs
that are not identified by the transitive closure algorithm.
Why is this?  Certainly, some state pairs $(s, t)$
introduced by the first type of
transformation have macro labels that,
 if executed one operator at a time, would stay within
the set $S$, and hence are pairs that are discovered by the
transitive closure algorithm.  
However, the second type of transformation may apply such a macro
to other states to discover pairs $(s, t) \in S \times S$ 
that would \emph{not}
be discovered by the transitive closure: this occurs when
a step-by-step execution
of the macro, starting from $s$, would leave the set $S$ 
before arriving to $t$.
Indeed, these two transformations depend on and feed off of each other:
the first transformation introduces increasingly powerful macros,
which in turn can be used by the second to increase the set of pairs,
which in turn permits the first to derive yet more powerful macros,
and so forth.

We now describe two concrete results to offer the reader a feel for
the power of our algorithm.
Let $s$ be any state of a \emph{Blocksworld-arm} instance,
and let $S$ be the set $H(s,4)$ of states 
within Hamming distance $4$ of $s$.\footnote{
The Hamming distance between two states is defined as the number of
variables at which they differ.
}
Let us use the term \emph{subtower} 
to refer to a sequence of blocks
stacked on top of one another such that the top is clear.
We prove that our algorithm, given the set $S$, will discover macros
that move any subtower of $s$ onto the ground
(preserving the subtower structure).
As another result, let $s$ be the initial state of
the \emph{Towers of Hanoi} problem, for any number of discs;
and, let $S$ be the set $H(s,7)$ of states 
within Hamming distance $7$ of $s$.
We prove that our algorithm, given the set $S$, will 
discover macros that, starting from the state $s$, 
move any subtower of discs from the initial peg to either of the other pegs.
In particular, our algorithm will report that the goal state
is reachable from the initial state $s$.
Note that, in the case of \emph{Blocksworld-arm}, the constant $4$
is independent of the state $s$, and in particular is independent of
the height of subtowers; likewise, in \emph{Towers of Hanoi},
the constant $7$ is independent of the number of discs.
Note also that, as can be proved, the transitive closure algorithm
does not detect either of these reachability conditions, even when
$S = H(s,k)$ for an arbitrarily large constant $k$.\footnote{
In the case of \emph{Towers of Hanoi}, this follows immediately from
the known exponential lower bound on the length of a plan transforming
the initial state to the goal state.  For a fixed $k \geq 1$,
when given the initial state and $H(s,k)$,
the transitive closure algorithm ``stays within the set'' $H(s,k)$,
which is of polynomial $O(n^k)$ size, and will not discover pairs
$(v, v')$ which are not linked by polynomial length plans.
}
We emphasize again that our new algorithm is fully domain-independent.

Our algorithm not only returns pairs of states, but also returns,
for each state pair $(s, t)$, a succinct representation of a plan 
from $s$ to $t$, as in~\cite{jonsson07}.  Note that our algorithm
may discover pairs $(s, t)$ for which the shortest plan from
$s$ to $t$ is of exponential length, when measured in terms of the 
original operators, as in the \emph{Towers of Hanoi} domain.

\paragraph{\bf Towards a tractability theory of domain-independent planning.}
Many of the benchmark domains--such as \emph{Blocksworld-arm}, \emph{Gripper},
and \emph{Logistics}--can now be handled
effectively and simultaneously by domain-independent planners,
as borne out by empirical evidence~\cite{hn01}.
This \emph{empirically observed} domain-independent tractability
of many common benchmark domains naturally calls for a
\emph{theoretical explanation}.  
By a theoretical explanation, we mean the formal
definition of tractable classes of planning instances, and
formal proofs that domains of interest fall into the classes.
Clearly, such an explanation could bring to the fore
structural properties shared by these benchmark domains.

To the best of our knowledge, research proposing tractable
classes has generally had other foci, such as
understanding syntactic restrictions on the operator 
set~\cite{bylander94,bn95,ens95},
studying restrictions of the causal graph, 
as in~\cite{bd03-unary,bd06,helmert06-fd,jonsson07},
or empirical evaluation of simplification rules~\cite{haslum07}.
Aligned with the present aims is
 the work of Hoffmann~\shortcite{hoffmann-utilizing} that
gives proofs that certain benchmark domains are solvable by local search
with respect to various heuristics.

To demonstrate the efficacy of our algorithm, we use it to extend
previously defined tractable classes.
In particular, previous work~\cite{act-local}
 presented a complexity measure called
\emph{persistent Hamming width (PH width)}, and demonstrated that
any set of instances having bounded PH width--PH width $k$ for some
constant $k$--is
polynomial-time tractable.
It was shown that both 
the \emph{Gripper} and \emph{Logistics} domains have bounded
PH width, giving a uniform explanation for their tractability.
In the present paper, we show that an extension of this measure
yields a tractable class containing
both the \emph{Blocksworld-arm} and \emph{Towers of Hanoi} domains,
and we therefore obtain a single tractable class which captures
all four of these domains.  
As mentioned, we believe that this is significant as
theoretical treatments have generally had limited coverage of
construction-type
domains such as \emph{Blocksworld-arm} and \emph{Towers of Hanoi}.  

We want to emphasize that our objective here is \emph{not}
to simply establish tractability of the domains under discussion:
in them, plan generation is already well-known to
be tractable on an individual, domain-dependent basis.
Rather, 
our objective is to give a \emph{uniform}, \emph{domain-independent} 
explanation 
for the tractability of these domains.  
Neither is our goal to prove that these domains have low time complexity;
again, our primary goal is to present a simple, domain-independent
algorithm for which we can establish 
tractability of these domains with respect to the heavily-studied
and mathematically robust concept of polynomial time.

\paragraph{\bf Previous work on macros.}
Macros have long been studied in planning~\cite{fn71}.
Early work includes \cite{minton85}, which developed filtering 
algorithms for discovered macros, and 
\cite{korf85}, which demonstrated the ability of macros to
exponentially reduce the size of the search space.

Macros have been thoroughly applied in domain-specific scenarios
such as puzzles and other games.  To name some examples,
there has been work on the sliding tile puzzle~\cite{iba89},
Sokoban~\cite{js01}, and Rubik's cube~\cite{h01}.

Some recent research on integrating macros into domain-independent
planning systems is as follows.  
\emph{Macro-FF}~\cite{bems05} is an extension of FF that has the ability
to automatically learn and make use of macro-actions.
\emph{Marvin}~\cite{cs07} is
a heuristic search planner 
that can form so-called macro-actions
upon escaping from plateaus that can be reused for future escapes.
Both of these planners participated in the 
International Planning Competition (IPC).
A method for learning macros given an arbitrary planner and example
problems from a domain is given in~\cite{nlfl07}.

A more theoretical approach was taken by~\cite{jonsson07},
who studied the use of macros in conjunction with causal graphs.
This work gives tractability results, and in particular shows that
domain-independent planners can cope with exponentially long plans
in polynomial time, which is also a feature of the present work.

The use of macros in this paper contrasts with that of most works
in that macros are generated and applied not over a domain or
even over an instance, but 
with respect to a ``current state'' $s$ and
a (small) set of related states $S$.  This ensures that the
 macros generated
are tailored to the state set $S$, and no filtering 
due to over-generation of macros is necessary.



\section{Preliminaries}

An instance of the planning problem is a tuple 
$\Pi = (V, \init, \goal, A)$ 
whose components are described as follows.

\begin{itemize}

\item $V$ is a finite set of variables, where each 
variable $v \in V$ has an associated
finite domain $D(v)$.
Note that variables are not necessarily propositional, that is,
$D(v)$ may have any finite size.
A \emph{state} is a mapping $s$ defined on the variables $V$
such that $s(v) \in D(v)$ for all $v \in V$.
A \emph{partial state} is a mapping $p$ defined on a subset 
$\vars(p)$ of the variables $V$ such that for all $v \in \vars(p)$,
it holds that $p(v) \in D(v)$.

\item $\init$ is a state called the \emph{initial state}.

\item $\goal$ is a partial state.

\item $A$ is a set of \emph{actions}.
An action
$a$ consists of a 
\emph{precondition} $\pre(a)$, which is a partial state,
as well as a \emph{postcondition} $\post(a)$, also a partial state.
We sometimes denote an action $a$ by $\langle \pre(a); \post(a) \rangle$.

\end{itemize}
Note that when $s$ is a state or partial state, 
and $W$ is a subset of the variable set $V$,
we will use $(s \restrict W)$ to denote the partial state resulting from
restricting $s$ to $W$.
We say that a state $s$ is a \emph{goal state} if
$(s \restrict \vars(\goal)) = \goal$.

We say that an action $a$ is \emph{applicable} at a state $s$
if 
$(s \restrict \vars(\pre(a))) = \pre(a)$.
We define a \emph{plan} to be 
a sequence of actions $P = a_1, \ldots, a_n$.
We will always speak of actions and plans
 relative to some planning instance $\Pi = (V, \init, \goal, A)$,
but we want to emphasize that when speaking (for example) of an action,
the action need not be an element of $A$; we require only that
its precondition and postcondition are partial states over $\Pi$.

Starting from a state $s$, we define the state resulting from $s$
by applying a plan $P$, denoted by $s[P]$, inductively as follows.
For the empty plan $P = \epsilon$, we define $s[\epsilon] = s$.
For non-empty plans $P$, denoting $P = P' , a$, we define
$s[P' , a]$ as follows.
\begin{itemize}

\item If $a$ is applicable at $s[P']$, then $s[P', a]$ is
 the state equal to $\post(a)$ on variables $v \in \vars(\post(a))$,
and equal to $s[P']$ on variables $v \in V \setminus \vars(\post(a))$.

\item Otherwise,  $s[P', a] = s[P']$.

\end{itemize}
We say that a state $s$ is \emph{reachable} (in an instance $\Pi$)
if there exists a plan $P$ such that $s = \init[P]$.
We are concerned with the problem of \emph{plan generation}:
 given an instance 
$\Pi = (V, \init, \goal, A)$ obtain a plan $P$ that \emph{solves} it,
that is, a plan $P$ such that $\init[P]$ is a goal state.

Note that sometimes we will use the representation of a partial function
$f$
as the relation
$\{ (a, b): f(a) = b \}$.

\section{Macro Computation Algorithm}

In this section, we develop
our macro computation algorithm.
This algorithm makes use of
a number of algorithmic subroutines.
In particular, we will present the two macro-producing operations
discussed in the introduction, $\apply$ and $\transitive$.
First, we define the notion of \emph{action graph}, the data structure
on which these operations work.

\begin{definition}
An \emph{action graph} is a directed graph $G$
whose vertex set, denoted by $V(G)$, is a set of states,
and whose edge set, denoted by $E(G)$, consists of labelled edges
that are actions;
we denote the label of an edge $e$ by $l_G(e)$ 
(or $l(e)$ when $G$ is clear from context).
Note that for every ordered pair of vertices $(s, s')$, there may be
at most one edge $(s, s')$ in $E(G)$,\footnote{
That is, an action graph is not a multigraph.}
and each edge has exactly one label.
\end{definition}

We now define three functions which will themselves be used as
subroutines in $\apply$ and $\transitive$.

\begin{definition}
We define the algorithmic function 
$\better(a, (s, s'), G)$
as follows.
Type-wise, the function 
$\better(a, (s, s'), G)$ requires that
$a$ is an action, 
$G$ is an action graph, and $s$ and $s'$ are vertices in $G$.
The pseudocode for $\better(a, (s, s'), G)$ is as follows:

\begin{tiny}
\begin{verbatim}
better(a, (s, s'), G) returns boolean 
{
  if((s, s') not in E(G))
    return TRUE;

  if(pre(a) strictly contained in pre(l(s, s'))  AND
     post(a) contained in post(l(s, s')))
    return TRUE;  

  if(pre(a) contained in pre(l(s, s'))  AND
     post(a) strictly contained in post(l(s, s')))
    return TRUE;  

  return FALSE; 
}
\end{verbatim}
\end{tiny}
\end{definition}

\begin{definition}
We define the algorithmic function 
$\addlabel(G, s, s', a)$
as follows.
Type-wise, the function
$\addlabel(G, s, s', a)$ requires that
$G$ is an action graph, $s$ and $s'$ are vertices in $G$,
and $a$ is an action.
The pseudocode for $\addlabel(G, s, s', a)$ is as follows:

\begin{tiny}
\begin{verbatim}
addlabel(G, s, s', a) returns G'
{
  G' := G;
  if((s, s') not in E(G))
  {
    place (s, s') in E(G');
  }
  l_{G'}(s, s') := a; 
  return G';
}
\end{verbatim}
\end{tiny}
\end{definition}

We remark that in our pseudocode, the assignment operator $:=$
is intended to be a value copy 
(as opposed to a reference copy, as in some programming languages).

\begin{definition}
We define the algorithmic function 
$\combine(a, a')$ as follows.
Type-wise, the function
$\combine(a, a')$ requires that $a$ and $a'$ are actions.
We remark that in all cases where we use the function
$\combine(a, a')$, there will exist states $s_1, s_2$
such that $a$ is applicable at state $s_1$, 
$s_1[a] = s_2$, and $a'$ is applicable at state $s_2$.
The pseudocode for $\combine(a, a')$ is as follows:

\begin{tiny}
\begin{verbatim}
combine(a, a') returns action a''
{
  R := vars(pre(a))  setminus  vars(post(a));
  s := post(a)  union  (pre(a) | R);
  O := vars(post(a))  setminus  vars(post(a'));
  pr := pre(a)  union  (pre(a') - s);
  pos := post(a')  union  (post(a) | O);
  return <pr; pos setminus pr>;
}
\end{verbatim}
\end{tiny}

Here, the pipe symbol $|$ should be interpreted as function restriction, and
the subtraction symbol in $(\pre(a') - s)$
should be interpreted as a set difference, where
the partial functions $\pre(a')$ and $S$ are viewed as relations.
Intuitively,
the partial state $s$ represents what we know about a state
if all we are told is that the action $a$ has just been
successfully executed.
\end{definition}

\longversion{
\begin{example}
\label{ex:combine}
Let $b_1, b_2$ be distinct blocks in the Blocksworld-arm.  
We consider the computation of 
$\combine(\unstack_{b_1, b_2}, \putdown_{b_1})$.
We have $R = \emptyset$, since all variables in 
the precondition of $\unstack_{b_1, b_2}$ appear in the postcondition.
We have $s = \post(\unstack_{b_1, b_2})$.
And, we have $O = \{ \clear{b_2} \}$.
The returned value is thus
$\langle \clear{b_1} = \true, \on{b_1} = b_2, \arm = \empty;
         \on{b_1} = \btable, \clear{b_2} = \true \rangle$.
\end{example}
}

The following propositions identify key properties of the
$\combine$ function.

\begin{prop}
Let $a$, $a'$ be actions and let $s$ be a state.
The action $\combine(a, a')$ is applicable at $s$
if and only if $a$ is applicable at $s$ and $a'$ is applicable at $s[a]$.
When this occurs, $s[\combine(a, a')]$ is equal to 
$s[a, a']$.
\end{prop}

\begin{prop}
\label{prop:combine-associative}
The function $\combine$ is associative.
That is,
the action
$\combine(\combine(a_1, a_2), a_3)$ is equal to the action
$\combine(a_1, \combine(a_2, a_3))$,
assuming that there exists a state $s$ such that
$a_1$ is applicable in $s$, 
$a_2$ is applicable in $s[a_1]$, and
$a_3$ is applicable in $s[a_1, a_2]$.
\end{prop}

We may now define the promised macro-producing operations.

\begin{definition}
We define two algorithmic functions  
$\apply(G, A, a, s)$ and $\transitive(G, s_1, s_2, s_3)$.
Type-wise, the function
$\apply(G, A, a, s)$ requires that $G$ is an action graph,
$A$ is a set of actions, $a$ is an action, and $s$ is a vertex of $G$.
The pseudocode for $\apply(G, A, a, s)$ is as follows:

\begin{tiny}
\begin{verbatim}
apply(G, A, a, s) returns G'
{
  G' := G;
  if( a in A  OR  a appears as a label in G' )  {
    if( s[a] != s  AND  s[a] in V(G))  {
      if( better(a, (s, s[a]), G)  {
        G' := addlabel(G, s, s[a], a);
      }
    }
  }
  return G';
}
\end{verbatim}
\end{tiny}

Type-wise, the function $\transitive(G, s_1, s_2, s_3)$
requires that $G$ is an action graph, and that
$s_1$, $s_2$, and $s_3$ are vertices in $G$.
The pseudocode for $\transitive(G, s_1, s_2, s_3)$ is as follows.

\begin{tiny}
\begin{verbatim}
transitive(G, s_1, s_2, s_3) return G'
{
  G' := G;
  if((s_1, s_2) in E(G) and 
     (s_2, s_3) in E(G))  {
    a := l(s_1, s_2);
    a' := l(s_2, s_3);
    a'' := combine(a, a');
    if( better(a'', (s_1, s_3), G)  {
      G' := addlabel(G, s_1, s_3, a'');
    }
  }
  return G';
}
\end{verbatim}
\end{tiny}

Within the function $\transitive$, in the case that the
$\addlabel$ function is called and returns a graph
$G'$ that is different from the input graph $G$, we say that the
transition $(s_1, a'', s_3)$
(where $s_1, s_3, a''$ are the arguments passed to the $\addlabel$ function)
is \emph{produced} by the function.
\end{definition}

In general, we use the term \emph{transition} to refer to 
a triple $(s, a, s')$ consisting of states $s, s'$ and an action $a$
such that $a$ is applicable at $s$ and $s[a] = s'$.

\begin{definition}
An \emph{action graph program} over a set of states $S$ and a set of actions 
$A$ is a sequence of commands
$\Sigma = \sigma_1, \ldots, \sigma_n$
of the form
$\apply(G, A, a, s)$, with $s \in S$, 
or $\transitive(G, s_1, s_2, s_3)$, with $s_1, s_2, s_3 \in S$.
The execution of an action graph program takes place as follows.  
First, $G$ is initialized to be the action graph with $S$ as vertices
and no edges.
Then, the commands of $\Sigma$ are executed in order;
for each $i$, after $\sigma_i$ is executed, $G$ is replaced with the
returned value.
\end{definition}

\longversion{
\begin{example}
We give an example action graph program over 
the Blocksworld-arm domain with block set $B = \{ b_1, b_2 \}$.
The program is over the set of all actions $A$ with respect to this
block set, and the state set $S$ containing the following
three states:
$s_1$, where $b_1$ is on top of $b_2$;
$s_2$, where $b_1$ is in the arm and $b_2$ is on the table; and,
$s_3$ where both $b_1$ and $b_2$ are on the table.
Formally, we define
$s_1 = \{ \arm = \empty, \on{b_1} = b_2,    \clear{b_1} = \true, \on{b_2} = \btable, \clear{b_2} = \false \}$,
$s_2 = \{ \arm = b_1,    \on{b_1} = \arm,   \clear{b_1} = \false, \on{b_2} = \btable, \clear{b_2} = \true \}$, and
$s_3 = \{ \arm = \empty, \on{b_1} = \btable, \clear{b_1} = \true, \on{b_2} = \btable, \clear{b_2} = \true \}$.

Consider the program 
$\Sigma = \sigma_1, \sigma_2, \sigma_3, \sigma_4$ 
with four commands where
\begin{align*}
\sigma_1 & =  \apply(G, A, \unstack_{b_1, b_2}, s_1 )\\
\sigma_2 & =  \apply(G, A, \pickup_{b_2},       s_2 )\\
\sigma_3 & =  \apply(G, A, \putdown_{b_1},      s_2 )\\
\sigma_4 & =  \transitive(G, s_1, s_2, s_3).
\end{align*}
Let us consider the execution of the program.
Initially, $G$ is set to be the action graph with vertex set
$S = \{ s_1, s_2, s_3 \}$ and no edges.
Let us use $G_i$ to denote the graph returned by the command
$\sigma_i$.
After the execution of $\sigma_1$, we have
$G = G_1$ where $G_1$ is equal to $G$, but with a directed edge
from $s_1$ to $s_2$ with the action $\unstack_{b_1, b_2}$ as label.
When $\sigma_2$ executes, inside this command we have
$s[a] = s$ (since $s_2[\pickup_{b_2}] = s_2$) and thus the same
graph that was passed in is returned.
Thus, after $\sigma_2$, we have
$G = G_2 = G_1$.
The execution of $\sigma_3$ will produce a transition, and the
returned graph $G_3$ will be equal to $G_1 = G_2$, 
but with an edge from $s_2$ to $s_3$ with label $\putdown_{b_1}$.
Finally, the execution of $\sigma_4$ will also produce a transition,
and the returned graph $G_4$ will be equal to $G_3$,
but with an edge from $s_1$ to $s_3$ with label $a_c$ where
$a_c$ is the combined action from Example~\ref{ex:combine}.
\end{example}
}

The following is our macro computation algorithm.
As input, it takes a set of states $S$ and a set of actions $A$.
The running time can be bounded by
$O(n |S|^3( |A| + |S|^2 ))$, where $n$ denotes the number of variables.

\begin{tiny}
\begin{verbatim}
compute_macros(S, A) returns G, M
{
  M := empty;
  V(G) := S;
  E(G) := empty set;

  do  {
    A' := (A union l(E(G)));
    for all: a in A', s in V(G)  {
      G := apply(G, A, a, s);
    }

    for all s1, s2, s3 in V(G)  {
      G := transitive(G, s1, s2, s3);
      if(transitive produces a transition)  {
        append "l(s1, s3) = l(s1, s2), l(s2, s3)" to M;
      }
    }
  }
  while(some change was made to G)

  return (G, M);
}
\end{verbatim}
\end{tiny}

\paragraph{Understanding compute\_macros.}

By a \emph{combination} over $A$, we mean an action in $A$ or an action
that can be derived from actions in $A$ by (possibly multiple) applications
of the $\combine$ function.

\begin{definition}
We say that a transition $(s, a, s')$ 
is \emph{condition-minimal} with respect to a set of actions $A$ if for
any combination $a'$ over $A$, if $s[a'] = s'$ then
$\pre(a) \subseteq \pre(a')$ and $\post(a) \subseteq \post(a')$
(when $\pre(a)$, $\pre(a')$, $\post(a)$, and $\post(a')$
are viewed as relations).
\end{definition}

Having defined the notion of a \emph{condition-minimal} transition,
we can now naturally define the notion of a 
\emph{condition-minimal} program.

\begin{definition}
Relative to a planning instance $\Pi$, 
let $S$ be a set of states, and let $A$, $A'$ be sets of actions.
An \emph{$A$-condition-minimal-program} 
(for short, $A$-CM-program)
over states $S$ and actions $A'$
is an action graph program over $S$ and $A$
such that when executed,
$\apply$ is only passed pairs $(a, s)$ such that $(s, a, s[a])$
is condition-minimal with respect to $A$, and 
the $\transitive$ commands produce
only transitions that are condition-minimal with respect to $A$.
\end{definition}


We now define a notion of \emph{derivable} action.  This notion is 
defined recursively.  Roughly speaking, 
derivable actions are actions that will provably be discovered
as macros by the algorithm.

\begin{definition}
Relative to a planning instance $\Pi$, let $S$ be a set of states, and
let $A$ be a set of actions.  We define the set of $(S, A)$-derivable
actions recursively, as the smallest set satisfying: any action of a
transition produced by an $A$-CM-program over states $S$ and the set
of actions that are $(S, A)$-derivable or in $A$, is $(S, A)$-derivable.
\end{definition}

\begin{lemma}
\label{lemma:discovery}
Relative to a planning instance $\Pi$ with action set
$A$, let $s$ be a state.
Any $(H(s,k), A)$-derivable action is discovered
by a call to the function \verb1compute_macros1 with the first two
arguments $H(s, k)$ and $A$, by which we mean that any such an action
will appear as an edge label in the graph output by \verb1compute_macros1.
\end{lemma}

We emphasize that, in the \verb1compute_macros1 procedure,
labels of edges are merely actions, which (as defined) are
precondition-postcondition pairs that need not appear in the
original set of actions $A$.  When new edge labels are introduced,
they are always obtained from existing labels or from $A$
via the \verb1combine1 procedure, which permits the general
applicability of edge labels.

\begin{pf-sketch}
Let $\Sigma = \sigma_1, \ldots, \sigma_n$ be an 
$A$-CM-program over $H(s, k)$ and actions that are discovered by
\verb1compute_macros1, and let $H$ be the graph returned by
\verb1compute_macros1; we prove the result by induction.

We consider the execution of the program $\Sigma$ with graph $G$.
We prove by induction on $i \geq 1$ that after the command
$\sigma_i$ is executed and returns graph $G_i$,
 for every edge
$(s, s') \in E(G_i)$, it holds that
$(s, s') \in E(H)$ and $l_{G_i}(s, s') = l_H(s, s')$.

If $\sigma_i$ is an $\apply$ command (with arguments $s$ and $a$)
that effects a change in the graph, 
then the input action must be in $l(E(G_i))$.
The command $\sigma_i$ can be successfully applied
at $H$.
Since $H$ is a fixed point over all $\apply$ and $\transitive$
commands, the action $a$ passed to $\apply$ 
or one that is better
(according to the function \verb1better1) must appear in $H$
at $l_H(s, s[a])$.
By condition-minimality of $(s, a, s[a])$, 
we have that $a = l_H(s, s[a])$.

If $\sigma_i$ is a $\transitive$ command that produces a transition
$(s, a, s')$, then the actions $a'$ and $a''$ (from within the 
execution of the command), by induction hypothesis, appear in $H$.
Since $H$ is a fixed point over all $\apply$ and $\transitive$ commands,
the action $\combine(a, a')$ or one that is better
must appear in $H$ at $l_H(s, s')$.  By condition-minimality of
$(s, \combine(a, a'), s')$, we have that 
$\combine(a, a') = l_H(s, s')$.
\end{pf-sketch}

\section{Examples}

\paragraph{Blocksworld-arm.}
We will present results with respect to the following formulation of the
Blocksworld-arm domain, which is based strongly on the 
propositional STRIPS formulation.
We choose this 
formulation primarily to lighten the presentation, and remark that
it is straightforward to verify that our proofs and results apply to the
propositional formulation.
\longversion{
We give the description of the used formulation right now, as we will use it
throughout the paper to present examples of introduced notions.}

\begin{domain} (Blocksworld-arm domain)
We use a formulation of this domain where there is an arm.
Formally, in an instance $\Pi=(V, \init, \goal, A)$
of the Blocksworld-arm domain, there is a set of blocks $B$,
and the variable set $V$ is defined as 
$\{ \arm \} \cup \{ \on{b}: b \in B \} \cup \{ \clear{b}: b \in B \}$ where
$D(\arm) = \{ \empty \} \cup B$ and
for all $b \in B$,
$D(\on{b}) = \{ \btable, \arm \} \cup B$
and
$D(\clear{b}) = \{ \true, \false \}$.
The $\on{b}$ variable tells what the block $b$ is on top of, or whether
it is being held by the arm, and the $\clear{b}$ variable tells
whether or not the block $b$ is clear.

\oldshortversion{
There are four kinds of actions: 
$\pickup_b$, $\putdown_b$, $\unstack_{b,c}$ and $\stack_{b,c}$.
For example, 
$\forall b \in B$, $\pickup_b = 
\langle \clear{b} = \true, \on{b} = \btable, \arm = \empty;
        \clear{b} = \false, \on{b} = \arm, \arm = b \rangle$.
Also, 
$\forall b, c \in B$, $\unstack_{b, c} = 
\langle \clear{b} = \true, \on{b} = c, \arm = \empty;
        \clear{b} = \false, \on{b} = \arm, \arm = b, \clear{c} = \true \rangle$.\end{domain}
}

\proofsversion{
There are four kinds of actions.
\begin{itemize}

\item $\forall b \in B$, $\pickup_b = 
\langle \clear{b} = \true, \on{b} = \btable, \arm = \empty;
        \clear{b} = \false, \on{b} = \arm, \arm = b \rangle$

\item $\forall b \in B$, $\putdown_b = 
\langle \arm = b;
        \arm = \empty, \clear{b} = \true, \on{b} = \btable \rangle$

\item $\forall b, c \in B$, $\unstack_{b, c} = 
\langle \clear{b} = \true, \on{b} = c, \arm = \empty;
        \clear{b} = \false, \on{b} = \arm, \arm = b, \clear{c} = \true \rangle$

\item $\forall b, c \in B$, $\stack_{b, c} = 
\langle \arm = b, \clear{c} = \true;
        \arm = \empty, \clear{c} = \false, \clear{b} = \true, \on{b} = c \rangle$

\end{itemize}

\longversion{
We say that a state $s$ is \emph{consistent} if it satisfies the following restrictions.
\begin{itemize}
\item $\forall b'\in B, s(\clear{b'}) = \true \Rightarrow | \{b \in B| s(\on{b})=b'\} | = 0$.
\item $\forall b'\in B, s(\clear{b'}) = \false \Rightarrow | \{b \in B| s(\on{b})=b'\} | = 1$.
\item $\forall b\in B, s(\arm) = b \Leftrightarrow s(\on{b}) = \arm$.
\end{itemize}

We define the planning domain Blocksworld-arm as the set of those planning
instances $\Pi$ such that the goal state is total and the states
$\init$ and $\goal$ are consistent. With a little more effort our results
extend to the case where the $\goal$ may be an arbitrary partial state.

It is easy to prove that in any Blocksworld-arm planning instance a state
$s$ is reachable if and only if $s$ is consistent. In particular, all
planning instances with consistent $\goal$ state are solvable.
}
\end{domain}
}


\newcommand{\pbottom}[1]{\mathsf{bottom}(#1)}
\newcommand{\ptop}[1]{\mathsf{top}(#1)}

\longversion{
We show in this section that the planning domain Blocksworld-arm has
MPH width $10$. This implies that the domain-independent planner
described in the previous section solves any Blocksworld-arm instance
in polynomial time.
We should remark that we believe a closer analysis will yield a
lower value for the MPH width of Blocksworld-arm; our main focus here is
simply showing \emph{bounded} MPH width.
}




\longversion{We introduce the following definitions.}

\begin{definition}
Relative to an instance $\Pi$ of Blocksworld-arm and a reachable state
$s$ of $\Pi$,
a \emph{pile} $P$ of $s$ is a non-empty sequence of blocks $(b_1,
\ldots, b_k)$ such that $s(\on{b_i})=b_{i+1}$ for all $i\in[1,k-1]$.
The \emph{top} of the pile $P$ is the block
$\ptop{P}=b_1$, and the \emph{bottom} of the pile is
the block $\pbottom{P}=b_k$. The \emph{size} of $P$ is $|P|=k$.

A \emph{sub-tower} of $s$ is a pile $P$ such that
$s(\clear{\ptop{P}})=\true$; a tower is a sub-tower such that
$s(\on{\pbottom{P}})=\btable$.

We use the notation $P_{\geq}(b)$ (respectively,
$P_{>}(b)$, $P_{\leq}(b)$, $P_{<}(b)$) to denote the sub-tower
with bottom block $b$ (respectively, the sub-tower stacked on $b$,
and the piles supporting $b$, either including $b$ or not.)

\end{definition}

\begin{definition}
Let $\Pi$ be a planning instance of Blocksworld-arm. Let $P=(b_1, \ldots,
b_k)$ be a sequence of blocks, and $b$ and $b'$ two different blocks
not in $P$. Let $S$ be the partial state $\{\clear{b_1}=\true,
\arm=\empty, \on{b_1}=b_2, \ldots, \on{b_{k-1}}=b_{k}\}$. We define
several actions with $S$ as common precondition.
\begin{itemize}
\item The action $\subtowerToTable_{P, b} = \langle S, \on{b_k}=b;
\on{b_k}=\btable, \clear{b}=\true \rangle$ moves a sub-tower $P$ from a
block $b$ to the table.

\item The action $\subtowerToBlock_{P, b, b'} = \langle S, \on{b_k}=b,
\clear{b'}=\true; \on{b_k}=b', \clear{b}=\true, \clear{b'}=\false
\rangle$ moves a sub-tower $P$ from a block $b$ onto a block $b'$.

\item The action $\towerToBlock_{P, b'} = \langle S, \on{b_k}=\btable,
\clear{b'}=\true; \on{b_k}=b', \clear{b'}=\false \rangle$ moves a
tower $P$ onto a block $b'$.
\end{itemize}
\longversion{We remind the reader
 that these actions do not belong to the set of actions $A$ of $\Pi$.}
\end{definition}


\begin{theorem} \label{th:move-towers}
Let $\Pi$ be a planning instance of Blocksworld-arm, and let $s$ be a
reachable state with $s(\arm)=\empty$. 

\begin{itemize}
\item If $P$ is a sub-tower of $s$ and $s(\on{b_k})=b$, then
$\subtowerToTable_{P,b}$ is 
$(H(s,4),A)$-derivable.

\item If $P$ is a sub-tower of $s$, $s(\on{b_k})=b$ and 
$s(\clear{b'})=\true$, then $\subtowerToBlock_{P,b,b'}$ is
$(H(s,5),A)$-derivable.

\item If $P$ is a tower of $s$, $s(\on{b_k})=\btable$ and
$s(\clear{b'})=\true$, then $\towerToBlock_{P,b'}$ is
$(H(s,4),A)$-derivable.

\end{itemize}

\end{theorem}

\begin{pf-sketch}
The proof has two parts.
First, we show that the aforementioned
actions are condition-minimal. Then, we describe how to obtain an
$A$-CM-program that produces the actions inside $H(s,5)$.  We consider
the case $a=\subtowerToBlock_{P,b,b'}$; 
the remaining actions
admit similar proofs that only require Hamming distance $4$.

To prove condition-minimality of action $a$ we consider any
combination $C=(a_1, \ldots, a_t)$ of primitive actions from $A$ such
that $s[C]=s[a]$. We must show that the actions $\unstack_{b_1, b_2},
\ldots, \unstack_{b_k, b}, \stack_{b_k, b'}$ appear in $C$ in the
given relative order, and that no matter what are the remaining
actions of $C$, this already implies that $\pre(a)\subseteq \pre(C)$
and $\post(a)\subseteq \post(C)$. We remark that the proof is not
straight-forward, since $\pre(C)$ and $\post(C)$ are the result of
applying the $\combine$ subroutine to several actions not yet determined.

To prove that there exists an $A$-CM-program that produces actions
$\subtowerToTable$ and $\towerToBlock$ inside $H(s,4)$
we use a mutual induction; we omit the proof here.  We then use these
results for $\subtowerToBlock$, the proof for which we sketch here.
Precisely, we now show that $\subtowerToBlock_{P,b,b'}$ is
$(H(s,5),A)$-derivable.

When $|P|=1$, we derive $\subtowerToBlock_{P,b,b'}$ by combining
actions $a_1=\unstack_{b_1,b}$ and $a_2=\stack_{b_1, b'}$. The states
$s[a_1]$ and $s[a_1,a_2]$ differ from $s$ respectively $4$ and $3$
variables, so both states lie inside $H(s,5)$. When $|P|=k$, let
$P'=P_{>}(b_k)$ in state $s$.  We use the derivable actions
$a_1=\subtowerToTable_{P', b_k}$, $a_2=\unstack_{b_k, b}$,
$a_3=\stack_{b_k, b'}$ and $a_4=\towerToBlock_{P', b_k}$. It is easy
to check that the state $s[a_1, a_2, a_3]$ is the one that is furthest
from $s$, differing at the $5$ variables $\clear{b}$,
$\on{b_{k-1}}$, $\clear{b_k}$, $\on{b_k}$ and $\clear{b'}$.
\end{pf-sketch}

\paragraph{Towers of Hanoi.}
\newcommand{\move}{\mathsf{move}}
\shortversion{
We study the formulation 
of \emph{Towers of Hanoi}
where, for every disk $d$, a variable stores
the position (that is, the disk or the peg) the disk $d$ is on.
Formally, in an instance $\Pi=(V, \init, \goal, A)$ of the Towers of
Hanoi domain, there is an ordered set of disks $D=\{d_1, \ldots,
d_k\}$ and a partially ordered set of positions $P=D\cup\{p_1, p_2,
p_3\}$, where $d_i<p_j$ for every $i$ and $j$. The set of variables
$V$ is defined as $\{\on{d}: d \in D\} \cup \{\clear{x}: x \in P\}$,
where $D(\on{d})=P$ and $D(\clear{x})=\{\true, \false\}$.

The only actions in Towers of Hanoi are movement actions that move a
disk $d$ into a position $x$, provided that both $d$ and $p$ are clear and
$d<x$.
\begin{itemize}

\item $\forall d \in D$, $\forall x, x'\in P$, if $d<x$, then define
$\move_{d, x', x} = \langle \clear{d} = \true, \clear{x}=\true,
\on{d} = x'; \clear{x} = \false, \clear{x'} = \true, \on{d}=
x\rangle$
\end{itemize}

We define this planning domain as the set of those planning
instances $\Pi$ such that the $\init$ and $\goal$ are certain
predetermined total states. Namely, in both states $\init$ and $\goal$
it holds $\on{d_i}=d_{i+1}$ for all $i\in[1, \ldots, k-1]$,
$\clear{d_1}=\true$, $\clear{d_i}=\false$ for all $i\in[2,k]$
and $\clear{p_2}=\true$. They only differ in three variables:
$\init(\on{d_k})=p_1$, $\init(\clear{p_1})=false$ and
$\init(\clear{p_3})=\true$, but $\goal(\on{d_k})=p_3$,
$\goal(\clear{p_1})=\true$ and $\goal(\clear{p_3})=\false$.

\begin{definition}
Let $\Pi$ be a planning domain instance of Towers of Hanoi. Let $i$ be
an integer $i\in[1,k]$. Let $x=\init(\on{d_i})$ and $x'\in\{p_2,
p_3\}$.  We define the action $\subtowerToPosition_{i, x, x'} =
\langle \clear{d_1}=\true, \on{d_1}=d_2, \ldots, \on{d_{i-1}}=d_i,
\on{d_i}=x, \clear{x'}=\true; \on{d_i}=x', \clear{x}=\true,
\clear{x'}=\false\rangle$, that is, the action that moves the tower
of depth $i$ 
from $x$ to $x'$.
\end{definition}

\begin{theorem}
The actions $\subtowerToPosition_{i, x, x'}$ are
$(H(\init,7),A)$-derivable.
\end{theorem}
}

We prove this by induction on $i$, the height of the subtower.
To derive actions of the form $\subtowerToPosition_{i+1, x, x'}$ from
the actions of the form $\subtowerToPosition_{i, x, x'}$,
we make use of the classical recursive solution to Towers of Hanoi;
an analysis shows that this recursive step stays within Hamming distance
$7$ of the initial state.

\longversion{
\begin{domain} (Towers of Hanoi domain)
We study the formulation where, for every disk $d$, a variable stores
the position (that is, the disk or the peg) the disk $d$ is on.
Formally, in an instance $\Pi=(V, \init, \goal, A)$ of the Towers of
Hanoi domain, there is an ordered set of disks $D=\{d_1, \ldots,
d_k\}$ and a partially ordered set of positions $P=D\cup\{p_1, p_2,
p_3\}$, where $d_i<p_j$ for every $i$ and $j$. The set of variables
$V$ is defined as $\{\on{d}: d \in D\} \cup \{\clear{x}: x \in P\}$,
where $D(\on{d})=P$ and $D(\clear{x})=\{\true, \false\}$.

The only actions in Towers of Hanoi are movement actions that move a
disk $d$ into a position $x$, provided that both $d$ and $p$ are clear and
$d<x$.
\begin{itemize}

\item $\forall d \in D$, $\forall x, x'\in P$, if $d<x$, then define
$\move_{d, x', x} = \langle \clear{d} = \true, \clear{x}=\true,
\on{d} = x'; \clear{x} = \false, \clear{x'} = \true, \on{d}=
x\rangle$
\end{itemize}

We define this planning domain as the set of those planning
instances $\Pi$ such that the $\init$ and $\goal$ are certain
predetermined total states. Namely, in both states $\init$ and $\goal$
it holds $\on{d_i}=d_{i+1}$ for all $i\in[1, \ldots, k-1]$,
$\clear{d_1}=\true$, $\clear{d_i}=\false$ for all $i\in[2,k]$
and $\clear{p_2}=\true$. They only differ in three variables:
$\init(\on{d_k})=p_1$, $\init(\clear{p_1})=false$ and
$\init(\clear{p_3})=\true$, but $\goal(\on{d_k})=p_3$,
$\goal(\clear{p_1})=\true$ and $\goal(\clear{p_3})=\false$.
\end{domain}


We show that the following actions are $(H(\init,7),A)$-derivable.

\begin{definition}
Let $\Pi$ be a planning domain instance of Towers of Hanoi. Let $i$ be
an integer $i\in[1,k]$. Let $x=\init(\on{d_i})$ and $x'\in\{p_2,
p_3\}$.  We define the action $\subtowerToPosition_{i, x, x'} =
\langle \clear{d_1}=\true, \on{d_1}=d_2, \ldots, \on{d_{i-1}}=d_i,
\on{d_i}=x, \clear{x'}=\true; \on{d_i}=x', \clear{x}=\true,
\clear{x'}=\false\rangle$, that is, the action that moves the tower
of depth $i$ 
from $x$ to $x'$.
\end{definition}
}

\section{Width}

In this section, we present the definition of 
macro persistent Hamming width and present the width results on domains.
For a state $s$, we define $\wrong(s)$ to be the variables that are not
in the goal state, that is,
$\wrong(s) = \{ v \in \vars(\goal) ~|~ s(v) \neq \goal(v) \}$.

\begin{definition}
With respect to a planning instance $(V, \init, \goal, A)$,
we say that a state $s'$ is an improvement of a state $s$ if
\begin{itemize}

\item for all $v \in V$, if $v \in \vars(\goal)$ and
$s(v) = \goal(v)$, then $s'(v) = \goal(v)$; and,

\item there exists $u \in \vars(\goal)$
such that $u \in \wrong(s)$ and 
$s'(u) = \goal(u)$.

\end{itemize}
In this case, we say that such a variable $u$ is a variable being improved.
\end{definition}

\begin{definition}
\label{def:plan-improves-state}
With respect to a planning instance $(V, \init, \goal, A)$,
we say that a plan $P$ \emph{improves} a state $s$
if $s[P]$ is a goal state, or $s[P]$ is an improvement of $s$.
\end{definition}

\longversion{
We remark that in Definition~\ref{def:plan-improves-state}, we permit
$P$ to be the empty plan $\epsilon$; in particular,
we have that the empty plan improves any goal state.

\begin{definition}
A planning instance $(V, \init, \goal, A)$
has \emph{persistent Hamming width $k$}
(for short, \emph{PH width $k$}) if no plan exists, or 
for every reachable state $s$, 
there exists a plan (over $A$) improving $s$ that 
stays within Hamming distance $k$ of $s$.
\end{definition}

In this definition, when we say 
that a plan stays within Hamming distance $k$ of a state $s$,
we mean that when the plan is executed in $s$, all intermediate states
encountered (as well as the final state) are within Hamming distance $k$
of $s$.
As the empty plan improves any goal state, to show that a
planning instance has PH width $k$ (according to the given definition),
the interesting case is to consider reachable states $s$ that are 
not goal states.
}


Relative to a planning instance,
we say that a state $s$ dominates another state $s'$ if
$\{ v \in V: s(v) \neq s'(v) \} \subseteq \vars(\goal)$ and
$\wrong(s) \subseteq \wrong(s')$; intuitively, $s'$ may differ from $s$
only in that it may have more variables set to their goal position.
Recall that
for a state $s$ and natural number $k \geq 0$, 
we use $H(s, k)$ to denote the set of all states within Hamming distance
$k$ from $s$.

We now give the official definition of our new width notion.

\begin{definition}
A planning instance $(V, \init, \goal, A)$
has \emph{macro persistent Hamming width $k$}
(for short, \emph{MPH width $k$}) if no plan exists, or
for every reachable state $s$ dominating the initial state $\init$,
there exists a plan over $(H(s,k), A)$-derivable actions
improving $s$ that stays within Hamming distance $k$ of $s$.
\end{definition}

It is straightforwardly verified that if an instance has
PH width $k$, then it has MPH width $k$.

We now give a
polynomial-time algorithm for sets of planning instances
having bounded MPH width.  We establish the following theorem.

\begin{theorem}
\label{thm:mph-algorithm}
Let $\mathcal{C}$ be a set of planning instances having
MPH width $k$.  The plan generation problem for $\mathcal{C}$
is solvable in polynomial time via the following algorithm,
in time
$O(n^{3k+2} d^{3k} (a + (nd)^{2k}))$.
Here, $n$ denotes the number of variables,
$d$ denotes the maximum size of a domain,
and
$a$ denotes the number of actions.
\end{theorem}

\begin{tiny}
\begin{verbatim}
solve_mph((V, init, goal, A), k)
{
  Q := empty plan;
  M := empty set of macros;
  s := init;

  while( s not a goal state )  {
    (G, M') := compute_macros(H(s,k), A);
    append M' to M;    

    if(an improvement s' of s is reachable from s in G)  {
      s := s';
    }
    else  {
      print "?";
      halt;
    }
    append l(s, s') to Q;
  }
  print M;
  print Q;
}
\end{verbatim}
\end{tiny}

\longversion{By the notation $l(E(G))$, we mean the set of labels
$\{ l(e): e \in E(G) \}$.
We remark that \verb1solve_mph1 can really be viewed as an extension 
of an algorithm for persistent Hamming (PH) width; one essentially
obtains an algorithm for PH width from \verb1solve_mph1 by 
replacing the call to \verb1compute_macros1 with a command that 
simply sets $G$ to be the directed graph with vertex set $H(s, k)$ and
an edge $(s_1, s_2)$ present if there is an action $a$ in $A$
such that $s_1[a] = s_2$.
}



\begin{pf-sketch} 
Let $\Pi \in \mathcal{C}$ be a planning instance such that
there exists a plan for $\Pi = (V, \init, \goal, A)$.  We want to show that
\verb1solve_mph1 outputs a plan.  
During the execution of \verb1solve_mph1, the state $s$ can only
be replaced by states that are improvements of it,
and thus $s$ always dominates the initial state $\init$.
By definition of MPH width, then, for any $s$ encountered during
execution, there exists a plan over $(H(s, k), A)$-derivable actions
improving $s$ staying within Hamming distance $k$ of $s$.
By Lemma~\ref{lemma:discovery}, all of the actions are discovered by
\verb1compute_macros1, and thus the reachability check in 
\verb1solve_mph1 will find an improvement.

We now perform a running time analysis of the algorithm.
Let $v$ denote the number of vertices in the graphs in
\verb1compute_macros1, that is, $|H(s,k)|$.  
We have $v \leq {n \choose k} d^k \in O((nd)^k)$.
Let $e$ be the maximum number of edges; we have
$e = {v \choose 2} \in O((nd)^{2k})$.
The \verb1do-while1 loop in \verb1compute_macros1 will execute
at most $2n \cdot e \in O(ne)$ times, since once an edge is introduced,
its label may change at most $2n$ times, by definition of \emph{better}.
Each time this loop iterates, it uses no more than
$(a + e)v + v^3$ time: 
\emph{apply} can be called on no more than
$(a+e)v$ inputs, and \emph{transitive}
 can be called on no more than $v^3$ inputs.
The while loop in \verb1solve_mph1 loops at most $n$ times, 
and each time, by the previous discussion,
it requires $ne((a+e)v + v^3)$
time for the call to \verb1compute_macros1,
and $(v+e)$ time for the reachability check.
The total time is thus
$O(n( ne((a+e)v + v^3) + (v + e)))$
which is
$O(n^2 e((a+e)v + v^3))$
which is
$O(n^2 e(a+e)v)$ which is
$O(n^{3k+2} d^{3k} (a + (nd)^{2k}))$.
\end{pf-sketch}

\shortversion{
\paragraph{Blocksworld.}

\begin{theorem}
\label{thm:blocksworld-10}
All instances of the Blocksworld-arm domain have MPH-width $10$.
\end{theorem}

\oldshortversion{
In the proof of this theorem, we show how to
improve any reachable state $s$ of \emph{Blocksworld-arm}.  
The proof is conceptually simple:
improve $s$ just by moving around a few piles of blocks. For instance,
if $s(\on{b})=b'$ but $\goal(\on{b})=b''$, apply actions
$\subtowerToTable_{P_{>}(b''), b''}$, $\subtowerToBlock_{P_{\geq}(b),
b', b''}$. There is, however, a technical difficulty:
 we must not forget that variables that were
already in the goal state in $s$ must remain so after the
improvement. For instance, if $b$ was on top of $b'$ in $s$, then
unstacking $b$ from $b'$ will make $\clear{b'}$ change from $\false$
to $\true$. 
The solution to this involves considering that
if something is placed on
top of $b'$, this movement may affect some other variable
which was already in the goal state, and so forth.
}
}

\proofsversion{

According to Theorem~\ref{th:move-towers}, at any state $s$ we may
consider our set of applicable actions enriched by this new macro-actions.
We now show how can these new actions be used to improve any reachable
state $s$. The proof is conceptually simple:
improve $s$ just by moving around a few piles of blocks. For instance,
if $s(\on{b})=b'$ but $\goal(\on{b})=b''$, apply actions
$\subtowerToTable_{P_{>}(b''), b''}$, $\subtowerToBlock_{P_{\geq}(b),
b', b''}$. However, we must not forget that variables that were
already in the goal state in $s$ must remain so after the
improvement. For instance, if $b$ was on top of $b'$ in $s$, then
unstacking $b$ from $b'$ will make $\clear{b'}$ change from $\false$
to $\true$. We may try to solve this by placing anything whatever on
top of $b'$, but then this movement may affect some other variable
which was already in the goal state, and so forth.

The following lemma is a case-by-case analysis of the solution to the
difficulty we have described.

\begin{lemma} \label{lem:clear-is-improvable}
Let $\Pi$ be an instance of the Blocksworld-arm domain, and let $s$ be a
reachable state of $\Pi$ such that $s(\arm)=\empty$. If a block $b$ is
such that $s(\clear{b})=\true$ but $\goal(\clear{b})=\false$, then
there is a plan using $(H(s,6),A)$-derivable actions that improves the
variable $\clear{b}$ in $s$.
\end{lemma}

\begin{pf-sketch}
Clearly, $b=\ptop{P_1}$ for some tower $P_1$ of $s$. Let $P_2, \ldots,
P_t$ be the remaining $t-1$ towers of $s$, and let $t'$ be the number
of towers of $\goal$.

The proof proceeds by cases.  If there is $i$ such that
$\goal(\on{\pbottom{P_i}}) \neq \btable$, we say we are in
Case~1. Otherwise, it holds that $t\leq t'$. In particular, there are
$t'$ blocks $b'$ such that $\goal(\clear{b'})=\true$ (block $b$ not
one of them), and $t$ blocks $b'\neq b$ such that
$s(\clear{b'})=\true$ (block $b$ being one of them). It follows that
it exists a block $b'$ such that $\goal(\clear{b'})=\true$ but
$s(\clear{b'})=\false$. We say we are in Case~2 if the block $b'$
belongs to the tower $P_1$, and in Case~3 if not.  Throughout this
proof we say that a block $b'$ is badly placed if
$s(\on{b'})\neq\goal(\on{b'})$.

\textbf{Case 1}. The tower $P_i$ is wrongly placed in the table, so we
are allowed to change the value of $\on{\pbottom{P_i}}$ without worry.

\begin{itemize}
\item[(a)] If $i\neq 1$, then use $\towerToBlock_{P_i, b}$ to stack the tower
$P_i$ on top of $b$. 

\item[(b)] If $i=1$ and a tower $P_j$ with $j>1$ has a badly
placed block $b'$, then a possible solution is to insert $P_1$ below
$b'$. That is, move the sub-tower $P_{\geq}(b')$ on top of $P_1$, and
then move the new resulting tower on top of the place where $b'$ was
in state $s$, that is, on top of $s(\on{b'})$.

\item[(c)] If $i=1$ and no tower $P_j$ with $j>1$ has badly placed blocks., then
consider the pile $P_i'$ in state $\goal$ that $b$ belongs to, and let
$b'=\top(P_i')$. If block $b'$ is in $P_j$ for $j>1$ in state $s$,
then $P_j$ would have some badly placed block, since $b'$ and $b$,
sharing pile $P_i'$ in the goal state, would be in different piles in
state $s$. So $b'$ is in $P_1$, $\goal(\clear{b'})=\true$ but
$s(\clear{b'})=\false$, since $b$ is the top of $P_1$. It follows that
the block on top of $b'$ in pile $P_1$ is badly placed. To improve
$\clear{b}$ use actions $\subtowerToTable_{P_{>}(b'),b'}$ and
$\towerToBlock_{P_\leq(b'), b}$, that is, break the tower over block
$b'$ and swap the two parts.
\end{itemize}

Note that an action like $\towerToBlock_{P_\leq(b'), b}$ is not
derivable from $s$ since the pile $P_\leq(b')$ is not a subtower of
$s$, but it is derivable from $s'=s[\subtowerToTable_{P_{>}(b'),b'}]$,
a state within distance $2$ from $s$. This fact may increase the width
required to discover the derivable actions. In our case, a careful
examination reveals that Situation~(b) requires width $5$ and
Situation~(c) requires width $4$. 

\textbf{Case 2}. Note that if Case~1 does not apply then $t\leq
t'$. Let $b'$ be the highest block in $P_1$ such that
$s(\clear{b'})=\false$ but $\goal(\clear{b'})=\true$.

\begin{itemize}
\item[(a)]
If $t>1$ and a tower $P_j$ with $j>1$ has a badly placed block
$b''$, then we insert the pile $P_{>}(b')$ below $b''$, analogously to
Situation~(b) in Case~1. This procedure improves variables $\clear{b}$ and
$\clear{b'}$ at the same time, but it needs width $6$.

\item[(b)]
If there is a second block $b''$ in $P_1$ such that
$\goal(\clear{b''})=\true$, then swap the sub-tower $P_{>}(b')$ with the pile
between $b'$ and $b''$, the block $b''$ not including. The procedure is
similar to Situation~(c) in Case~1, but it requires width 5.
 
\item[(c)] If there is no second block $b''$ in $P_1$ but all the
towers $P_j$ with $j>1$ have no badly placed blocks, it follows that
either $t=1$ or all towers $P_j$ with $j>1$ are exactly as in the goal
state. Observe that, in this situation, the blocks of $P_1$ form a
tower in $s$ and in $\goal$, but the order of the blocks in the two
towers must differ: the pile $P'=P_{\leq}(b')$, which is such that
$\goal(\clear{\ptop{P'}})=\true$ and
$\goal(\on{\pbottom{P'}})=\btable$, cannot be a pile in $\goal$. Hence
there is a badly placed block below $b'$. This situation is analogous
to Situation~(b) in Case~2, and it also requires width 5.
\end{itemize}

\textbf{Case 3}. There is a block $b'$ such that $s(\clear{b'})=\false$
but $\goal(\clear{b'})=\true$, and the block is in some tower $P_i$ other than
$P_1$. We just stack the sub-tower $P_{>}(b')$ on top of $b$.
\end{pf-sketch}

\begin{pf-sketch}
(of Theorem~\ref{thm:blocksworld-10})
Let $\Pi$ be an instance of the Blocksworld-arm domain, and let $s$ be a
reachable state of $\Pi$ that is not a goal state.  
We present the case where
$s(\arm)=\goal(\arm)=\empty$. 
\longversion{
Assume for the
moment that $s(\arm)=\goal(\arm)=\empty$. 
The proof is by cases,
depending on the variable to improve.}

\textbf{Improving $\on{b}$}.
\begin{itemize} 
\item $s(\on{b})=\btable, \goal(\on{b})=b'$. If
  $s(\clear{b'})=\false$, then move the sub-tower $P_{>}(b')$ onto the
  table. (This changes the variable $\on{b''}$, where $b''$ is the
  block on top of $b'$ in $s$, which was not in the goal state in
  $s$.)  Now the block $b'$ is clear, so we stack the tower $b$ is the
  bottom of onto $b'$.

\item $s(\on{b})=b'', \goal(\on{b})=b'$. If $s(\clear{b'})=\false$
  then we can swap piles $P_{>}(b'')$ and $P_{>}(b')$. Otherwise, we
  stack $P_{>}(b'')$ on top of $b'$, but then $\clear{b''}$ becomes
  true. This is a problem if $\goal(\clear{b''})=\false$, so we may
  need to apply Lemma~\ref{lem:clear-is-improvable} at the current
  state. Again, a careful examination shows that we may need width
  $8$.

\item $s(\on{b})=b'', \goal(\on{b})=\btable$. Move $P_{\geq}(b)$ onto
  the table. As in the previous case apply
  Lemma~\ref{lem:clear-is-improvable} to the current state if
  $\goal(\clear{b''})=\false$. In this case we may need width 7.
\end{itemize}

\textbf{Improving $\clear{b}$}.
\begin{itemize}
\item $s(\clear{b})=\false, \goal(\clear{b})=\true$. Move the pile $P_{>}(b)$
  onto the table, so width $4$ is enough.

\item $s(\clear{b})=\true, \goal(\clear{b})=\false$. Just apply
  Lemma~\ref{lem:clear-is-improvable}, which requires width 6.
\end{itemize}

Under the assumption that $s(\arm)=\goal(\arm)=\empty$, there is nothing
else to show, since we have explained how to improve any variable.
The width number $10$ comes from the analysis of the other cases.
\longversion{
Now, if the arm holds some block $b$ and $s(\arm)=b$ but
$\goal(\arm)\neq b'$, we can just place the block $b$ in the
table. The values of the variables $\clear{b}$, $\on{b}$ and
$\arm$ change, but they were not in the goal state, with the possible
exception of $\clear{b}$. (In the formulation we have given the
variable $\clear{b}$ is $\false$ when the arm holds the block $b$.)

We show how to improve in state $s$ a variable $v$ other than $\arm$,
$\on{b'}$, $\clear{b'}$. Let $s'$ be the state after dropping the
block $b$ on the table. Note that $s(\clear{b})=\false$ but
$s'(\clear{b})=\true$; if $b$ is such that $\goal(\clear{b})=\false$
then we have to improve both $v$ and $\clear{b}$. (The other variables
that change when dropping $b$, namely $\arm$ and $\on{b}$, could not
be in their goal state in $s$).  Now $s'(\arm)=\empty$, so we apply
the procedures to improve $v$ and $\clear{b}$ developed in the first
part of the proof. Observe that $s'$ differs from $s$ at $3$
variables; we have shown that $v$ can be improved with width $11$, and
that $\clear{b}$ needs at most width $8$. The sum of these values is a
trivial upper bound to the width required. (The actual minimal width
is certainly smaller, since the two improvements may modify common variables.)

If $v$ is either $\arm$, $\on{b'}$ or $\clear{b'}$, then we also drop
$b$ on the table and improve $\clear{b}$ if needed, but then we apply
procedures that make the block $b'$ become clear and on the table. If
$b'$ was in top of $b''$ we are changing $\clear{b''}$ from $\false$
to $\true$, so we may need to use Lemma~\ref{lem:clear-is-improvable}
a second time to make $\clear{b''}$ recover the value
$\false$. Finally, we pick up $b'$ with the hand, so that variables
$\arm$, $\on{b'}$ and $\clear{b'}$ improve at the same time. So during
the course of the plan we may need to improve variables $\clear{b}$
and $\clear{b''}$ (at most width $8$ for each one) and we are modifying
variables $\arm$, $\on{b}$, $\clear{b}$, $\on{b'}$ and $\clear{b'}$, so
we can certainly do it with width $21$.}
\end{pf-sketch}
}

\paragraph{Towers of Hanoi.}


\begin{theorem}
All instances of the Towers of Hanoi domain have MPH-width $7$.
\end{theorem}

Each instance can be solved by a single application of
the action $\subtowerToPosition_{k, p_1, p_3}$.

\begin{tiny}
\bibliographystyle{plain}
\bibliography{act}
\end{tiny}

\end{document}